\documentclass[letterpaper, 10 pt, conference]{ieeeconf}  

\IEEEoverridecommandlockouts                              

\usepackage{graphics} 
\usepackage{epsfig} 
\usepackage{mathptmx} 
\usepackage{times} 
\usepackage{amsmath} 
\usepackage{amssymb}  
\usepackage{booktabs} 
\usepackage{xcolor} 
\usepackage{tabularx}
\usepackage{cite}
\usepackage{hyperref} 

\title{\LARGE \bf
Direction-Scale Decomposition in Action Representation: Rethinking What to Tokenize for Vision-Language-Action Models}

\author{Yufei Duan$^{1}$, Hang Yin$^{2}$, Alberta Longhini$^{3}$, Chao Tang$^{1}$, Danica Kragic$^{1}$
\thanks{*Computational resources were provided on the Berzelius system funded by the Knut and Alice Wallenberg foundation and operated by NAISS. Computational resources were also supported on the Arrhenius system provided by the National Academic Infrastructure for Supercomputing in Sweden (NAISS), funded by the Swedish Research Council. We also acknowledge EuroHPC Joint Undertaking for awarding us access to Leonardo at CINECA, Italy. This work was partially supported by the Wallenberg AI, Autonomous Systems and Software Program (WASP) funded by the Knut and Alice Wallenberg Foundation.}
\thanks{$^{1}$Department of Robotics, Perception and Learning, KTH Royal Institude of Technology, Stockholm,
        {\tt\small \{yufeidu, chaotang, dani\}@kth.se}}%
\thanks{$^{2}$Department of Computer Science, University of Copenhagen, Copenhagen,
        {\tt\small \{hayi\}}@di.ku.dk}%
\thanks{$^{3}$Department of Computer Science, Stanford University, Stanford,
        {\tt\small \{alberta\}}@stanford.edu}
}

\begin{document}

\maketitle
\thispagestyle{empty}
\pagestyle{empty}

\begin{abstract}

Action representation plays a central role in discrete-token vision-language-action (VLA) learning but remains underexamined. Under conventional pose-increment representations, action tokens are sensitive to execution speed and dataset-specific normalization, potentially obscuring geometric structure shared across demonstrations and datasets. We introduce Direction-Scale Decomposition (DSD), an action representation that decomposes translation and rotation increments into direction and scale components before tokenization. DSD isolates motion direction while retaining magnitudes in separate scale channels. We evaluate DSD with uniform binning (BIN) and BEAST, a B-spline-based tokenizer, in simulation and real-world manipulation under both single-dataset and mixed-dataset training. On LIBERO, DSD improves average success rates with both tokenizers. On SimplerEnv, DSD-BIN outperforms BIN by 10.3 percentage points in overall success rate under mixed-dataset training. Real-robot experiments further show gains both with and without robotics pretraining. These results support DSD as an effective action representation for discrete-token VLA models and suggest its potential to mitigate performance degradation when training on large and diverse dataset mixtures. Our project page with additional resources is available at \url{https://vla-dsd.github.io/}.

\end{abstract}

\section{Introduction}
\label{sec:intro}

Vision-language-action (VLA) models have become a prominent approach to learning generalist robot policies~\cite{brohan2022rt1, zitkovich2023rt2, kim2024openvla, octo2024, black2024pi0}. One line of VLA research adopts the large language model (LLM) paradigm, formulating action prediction as autoregressive generation of discrete tokens. This formulation aligns action learning with the backbone's decoding architecture and cross-entropy objective, avoiding a separate action head whose competing training objective may interfere with pretrained vision-language capabilities~\cite{liang2025discretediffusionvla}. Robot demonstrations, however, are far scarcer than text and vary in both execution dynamics and numerical conventions, demanding greater care in the design of learning methods and system architectures.

\begin{figure}[t]
    \centering
    \includegraphics[width=\linewidth]{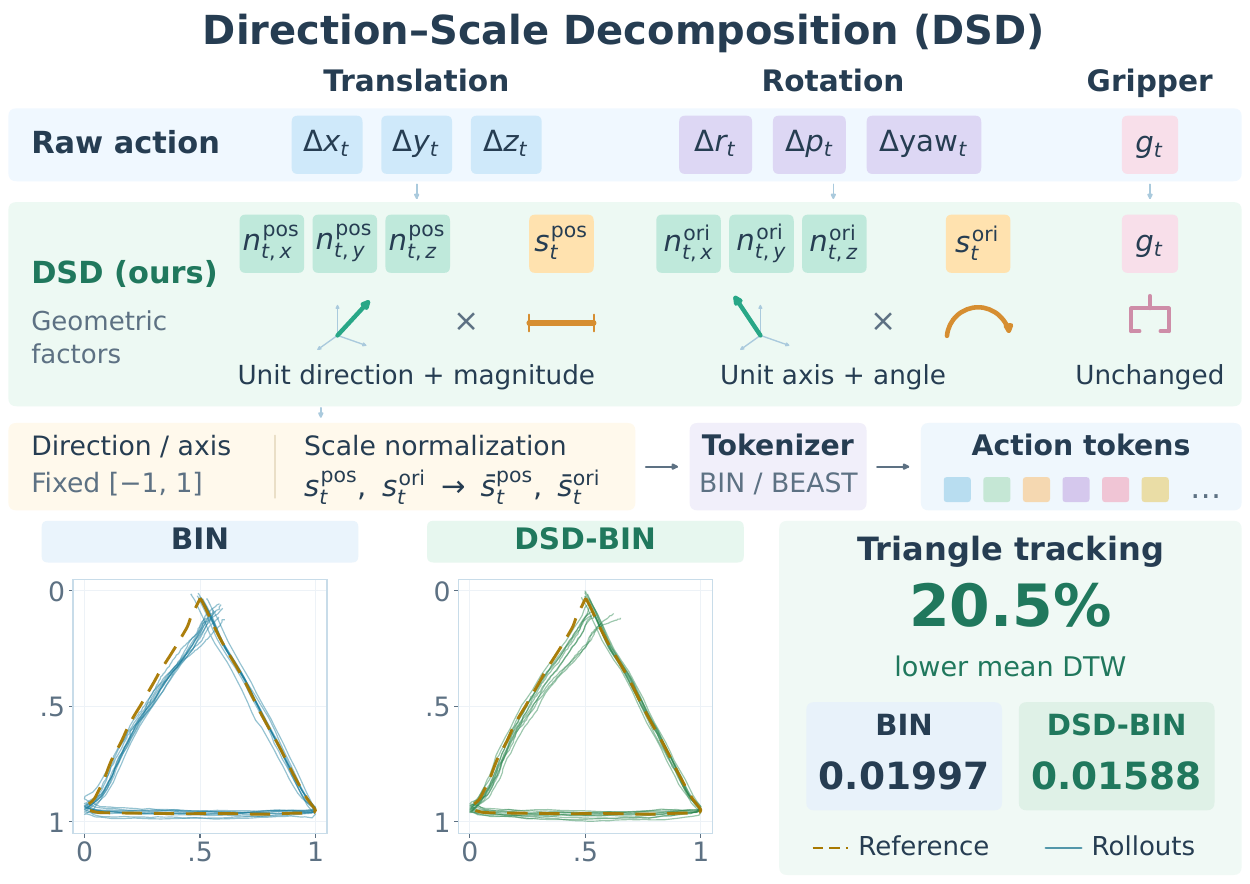}
    \caption{Direction--Scale Decomposition (DSD) separates translation into direction and magnitude and rotation into axis and angle before tokenization. DSD--BIN achieves 20.5\% lower mean DTW than BIN in triangle tracking.}
    \label{fig:intro}
\end{figure}

A common action representation is a delta end-effector pose ($\Delta$EEF) paired with a gripper command, which we refer to as the \emph{raw action representation}. It comprises seven dimensions per arm: three translation increments, three rotation increments, and one gripper command~\cite{brohan2022rt1, kim2024openvla, pertsch2025fast}. Before tokenization, these dimensions are typically normalized to $[-1,1]$ using dataset-specific statistics. Although often treated as a preprocessing step, normalization shapes the numerical structure presented to the tokenizer and thus the identities and patterns of the resulting action tokens.

We argue that this pipeline creates three problems. First, raw action representations entangle path geometry with task-irrelevant variation in execution speed: spatially similar trajectories executed at different speeds can map to dissimilar token sequences, obscuring task-relevant action patterns. Second, dataset-specific normalization can map the same physical action to different tokens across datasets, undermining cross-dataset learning. Third, coordinate-wise normalization can distort actions during transfer. When training and rollout statistics differ, denormalization can rescale the axes unequally and alter motion direction. This can turn an otherwise correct prediction into an incorrect or unsafe robot command. These failure modes are detailed in Section~\ref{sec:problem}.


While considerable attention has been devoted to action tokenization, the underlying raw action representation is often taken as given. We argue that revisiting this is important for addressing the challenges above. This motivates our main question: \emph{How can an action representation expose shared geometric structure across demonstrations despite variations in execution speed and dataset-specific normalization?}


Inspired by the complementary roles of direction and magnitude in word embeddings~\cite{wieting2015paraphrase}, we introduce \textbf{Direction--Scale Decomposition (DSD)}. As illustrated in Fig.~\ref{fig:intro}, DSD factorizes translation and rotation increments into direction--magnitude and axis--angle pairs. Directional components use fixed, dataset-independent bounds, while translation magnitudes and rotation angles are retained in separate scale channels. This separation preserves the magnitude information required for control while establishing a common numerical convention for directional components across datasets. DSD defines the geometric representation supplied to the tokenizer and can therefore be combined with different tokenization schemes.




Our contributions are twofold. First, we introduce DSD, an analytic action representation that factorizes translation and rotation increments into direction--magnitude and axis--angle pairs before scale normalization. This separation confines magnitude variation associated with execution speed to the scale channels while preserving directional semantics across scale calibrations. Second, we demonstrate DSD’s effectiveness in simulation and on real robots. On LIBERO, DSD improves average success rates under single-dataset training with both uniform binning and the B-spline-based BEAST tokenizer~\cite{zhou2025beast}. On SimplerEnv, DSD-BIN outperforms BIN by 10.3 percentage points under heterogeneous co-training. Improvement on path fidelity is also shown in a controlled path-tracking experiment with demonstrations collected at five speeds. Real-robot experiments demonstrate gains both without robotics pretraining and after fine-tuning pretrained checkpoints.

\section{Related Work}

\subsection{Vision-language-action models} 
VLAs such as RT-2~\cite{zitkovich2023rt2} and OpenVLA~\cite{kim2024openvla} adapt pretrained VLMs for action prediction, drawing on knowledge acquired from large-scale vision-language data. $\pi_0$~\cite{black2024pi0} uses flow matching, while $\pi_{0.5}$~\cite{pi2025pi05} combines discrete-token pretraining with flow-matching post-training. Despite differences in their action heads, these approaches generally use dataset-specific normalization. Our work examines how action representation interacts with motion scale and dataset-specific normalization in discrete-token VLAs.

\subsection{Action Tokenization}
\textbf{Uniform action binning.} Uniform binning is a simple and widely used strategy for action discretization. OpenVLA~\cite{kim2024openvla} and RT-2~\cite{zitkovich2023rt2} normalize each action dimension using dataset-specific statistics, discretize the resulting values into 256 uniform bins, and map bin indices to tokens in the language model's existing vocabulary. Although training-free, coordinate-wise binning of raw pose increments does not separate motion direction from magnitude. It therefore leaves the sensitivities to execution speed and normalization mismatch described in Problems~1--3 unresolved.



\textbf{Structured action tokenization.}
A second family of methods exploits the spatial-temporal structure to compress action chunks. FAST~\cite{pertsch2025fast} applies byte-pair encoding to quantized discrete cosine transform coefficients, while BEAST~\cite{zhou2025beast} fits B-splines to action chunks and quantizes their control points. Both improve token efficiency but operate on dataset-normalized actions. The coefficients and control points they tokenize therefore remain sensitive to execution speed and dataset-specific normalization.

\textbf{Learned action tokenization.} A third family learns a discrete codebook from data through vector quantization rather than relying on a fixed transformation. VQ-BeT~\cite{lee2024vqbet} uses residual vector quantization for individual actions or action chunks, while ActionCodec~\cite{dong2026actioncodec} develops codes that promote adjacent-chunk overlap, compactness, multimodal alignment, and reduced inter-token dependence. These methods, however, require a separately trained quantizer and may produce unreliable codes for action chunks outside the quantizer's training distribution. In contrast, DSD introduces no learned component and can be composed directly with both binning- and compression-based tokenizers.

\textbf{Geometry-inspired action tokenization.} Closer to our motivation, several action tokenizers explicitly exploit geometric or dynamical structure. Keypoint Action Tokens (KAT)~\cite{dipalo2024kat} represents absolute end-effector poses as triplets of 3D points, aligning actions with visual keypoints for in-context imitation. SpatialVLA~\cite{spatialvla} separates translational direction and distance using spherical coordinates and jointly discretizes them with adaptive action grids. MoMo~\cite{hu2026momo} learns separate spatial and temporal codebooks from joint trajectories and their dynamics for motion-mode conditioning and transfer. Nevertheless, KAT depends on absolute coordinate-frame conventions, SpatialVLA's tokens may remain sensitive to motion magnitude and normalization statistics, and MoMo intentionally retains speed-related variation in its temporal codes. By contrast, DSD is an action representation rather than a tokenizer: it analytically decomposes raw relative translation into direction and magnitude and relative rotation into axis and angle before normalization. Because DSD does not prescribe a particular tokenization scheme, it can provide structured inputs to, and complement, different downstream tokenizers.

\begin{figure*}[t]
    \centering
    \includegraphics[width=\linewidth]{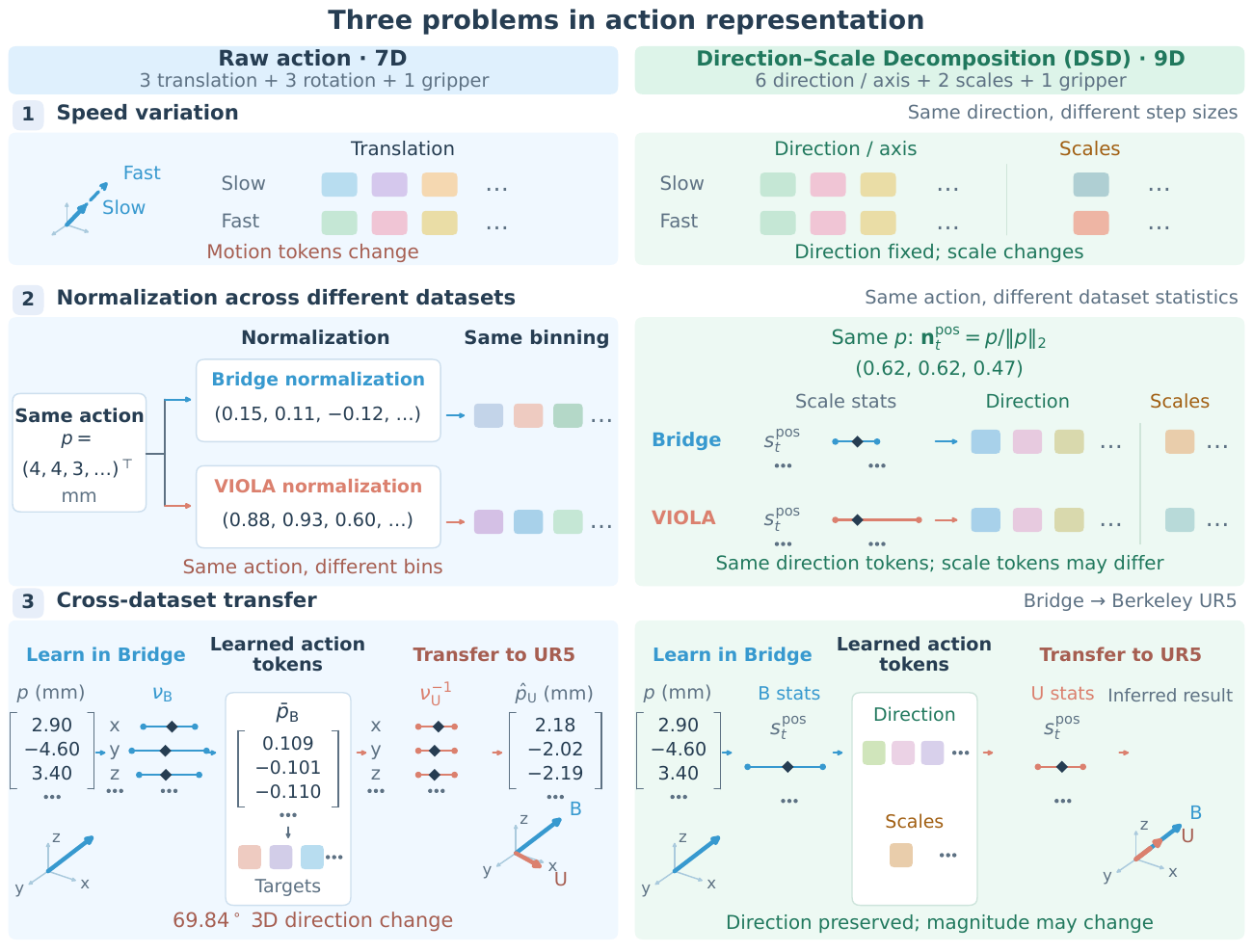}
    \caption{Direction–Scale Decomposition (DSD) preserves direction tokens under (1) speed variation and (2) dataset-specific normalization, while (3) preserving motion direction when decoding scales change. Token examples use shared per-coordinate binning.}
    \label{fig:placeholder}
\end{figure*}

\section{Preliminaries}

\textbf{Raw action representation and action space}
The \emph{raw action representation} parameterizes a per-arm delta end-effector (EEF) command before normalization and tokenization as $a_t=(\Delta x_t,\Delta y_t,\Delta z_t,\Delta r_t,\Delta p_t,\Delta yaw_t,g_t)^\top\in\mathcal{A}_{\mathrm{raw}}$, where $\mathcal{A}_{\mathrm{raw}}\subseteq\mathbb{R}^{3}\times\mathbb{R}^{3}\times\mathbb{B}$. The first six coordinates specify translation and roll--pitch--yaw rotation increments, and $\mathbb{B}\subset\mathbb{R}$ is the discrete set of gripper commands. Multi-arm commands are concatenated; we omit the arm index. An action chunk is $A_t=(a_t,\ldots,a_{t+T-1})$, where $T$ is the chunk length.

\textbf{Dataset-specific normalization}
In percentile-based normalization~\cite{kim2024openvla}, each coordinate is affinely mapped from its dataset's 1st and 99th percentiles, $q_{01,k}^{(i)}<q_{99,k}^{(i)}$, to $[-1,1]$ and clipped to this range. We denote this mapping by $\bar{a}_{t,k}=\nu_k(a_t)$. Dataset-specific statistics can therefore assign different normalized values to the same raw action.

\textbf{Tokenization and decoding}
An action tokenizer maps a normalized chunk to $Z_{t,k}=\mathcal{T}(\nu_k(A_t))\in\mathcal{V}^{*}$, where $\nu_k$ acts on each action and $\mathcal{V}^{*}$ denotes finite sequences over the action-token vocabulary $\mathcal{V}$. This formulation covers scalar binning, vector quantization, and chunk-level encoding. Uniform scalar binning assigns each normalized coordinate to one of $B$ equal-width bins on $[-1,1]$, producing $7T$ tokens per arm. OpenVLA~\cite{kim2024openvla} uses $B=256$ such tokens and predicts them through the standard language-model output head, which computes logits over the full vocabulary.

At inference, the policy predicts $\hat{Z}_t$ from visual observations $I_t$ and a language instruction $\ell$, then reconstructs commands as $\hat{A}_t=\nu_{k_\star}^{-1}(\mathcal{D}(\hat{Z}_t))$. Here, $\mathcal{D}$ returns normalized actions (bin centers for uniform binning), and $\nu_{k_\star}^{-1}$ applies inverse affine denormalization using the target dataset or robot setup's statistics. Thus, the physical commands associated with a token sequence depend on both the tokenizer and the denormalization statistics.

\section{Limitations of Current Action Representations}
\label{sec:problem}

Current action representations pose three problems for VLA learning, as illustrated in Fig.~\ref{fig:placeholder}. This section details how each problem can affect the model.

\subsection{Problem 1: irrelevant factors hide action patterns}

We use \emph{action patterns} to denote task-relevant geometric structure of an action, primarily its path geometry, rather than incidental numerical variation. The raw action representation, $\Delta$EEF, is directly tied to demonstration velocity. Paths that are spatially similar but executed at different speeds, a routine occurrence across teleoperators, and even across demonstrations from a single teleoperator, are assigned to widely separated tokens. Two demonstrations of the same task can therefore produce substantially different token sequences because of velocity variation, making task-irrelevant speed changes harder to distinguish from changes in behavior. Although the corresponding displacements may differ only by simple scale factors, categorical action tokens do not explicitly encode these relationships; the model must infer them indirectly from discrete token identities. With limited training data, speed-dependent numerical variants make reusable geometric patterns harder to learn, allowing task-irrelevant velocity variation to obscure their shared structure.

\subsection{Problem 2: Inconsistency in dataset normalization}

Dataset-specific normalization can assign different token targets to the same physical action, even when datasets share coordinate conventions and physical units. As illustrated in Fig.~\ref{fig:placeholder}, normalization using Bridge and VIOLA statistics maps an identical translation to different bins under the same 256-bin quantizer. This inconsistency originates in normalization rather than tokenization. Mixed-dataset training therefore requires the model to associate the same underlying motion with distinct token targets, hindering the reuse of shared action patterns across datasets. The difficulty of heterogeneous co-training has also been observed empirically: RT-1-X trained on the Open X-Embodiment dataset mixture performs worse than the RT-1 baseline trained only on Bridge~\cite{oxe2023}.

\subsection{Problem 3: direction distortion in action transfer} 


VLAs are often trained jointly on multiple datasets to promote skill transfer across tasks and embodiments. During training, action-token targets from each dataset are constructed using that dataset's normalization statistics. At rollout in a target setup, predicted tokens are decoded into normalized action values and then denormalized using the target statistics. If the model reuses a token pattern learned from a source dataset, this change in normalization can alter the resulting physical motion. The mismatch therefore affects more than token consistency: it can change the action direction even when the predicted token pattern is correct under the source normalization. Because each coordinate has a dataset-specific offset and scale, target-side denormalization can change the relative magnitudes or even reverse the signs of reconstructed motion components.

In the example shown in Fig.~\ref{fig:placeholder}, a positive $z$ displacement falls below the midpoint of Bridge's percentile range and therefore receives a negative normalized value. Denormalizing this value using Berkeley-UR5 statistics, whose midpoint is zero, produces a negative physical displacement. Thus, normalization mismatch can alter motion direction through both affine offsets and unequal axis scaling, even without clipping or quantization.

Using a single binning range shared across datasets and translation axes could avoid this mismatch, but the range must accommodate the largest motion scale. Datasets with smaller action ranges would then occupy only a narrow subset of the bins, leaving many bins unused and reducing effective quantization resolution.

\section{Decomposed scale and normalization}
\label{sec:method}

In this section, we introduce the Direction-Scale Decomposition and how it can be converted from the raw action representation. Lastly, we give some remarks about its implementation in the training procedure.

\subsection{Direction--scale action representation}

Direction--Scale Decomposition (DSD) separates translation into direction and magnitude and relative rotation into axis and angle. For each arm, the action at timestep $t$, before scale normalization and tokenization, is represented as
\begin{equation}
    \tilde{a_t} = 
    \bigl( s_t^{\mathrm{pos}}, s_t^{\mathrm{ori}}, g_t,
    (n_t^{\mathrm{pos}})^\top, (n_t^{\mathrm{ori}})^\top\bigr)^\top
    \in \mathcal A_{\mathrm{DSD}},
    \label{eq:dsd-representation}
\end{equation}

with the admissible action space satisfying

\begin{equation}
    \mathcal{A}_{\mathrm{DSD}}
    \subseteq \mathbb{R}^{2}\times\mathbb{B}
    \times\mathbb{S}^{2}\times\mathbb{S}^{2},
    \label{eq:dsd-action-space}
\end{equation}
where $\mathbb{B}\subset\mathbb{R}$ is the discrete set of admissible gripper commands and $\mathbb{S}^{2}=\{n\in\mathbb{R}^{3}:\|n\|_2=1\}$ is the unit sphere.

The scale coordinates $s_t^{\mathrm{pos}}$ and $s_t^{\mathrm{ori}}$ encode translation magnitude and rotation angle, respectively. The unit vectors $n_t^{\mathrm{pos}}$ and $n_t^{\mathrm{ori}}$ specify the translation direction and rotation axis, while $g_t\in\mathcal B$ retains the original gripper command.

Storing each unit vector as three Cartesian components gives nine scalar coordinates per arm while retaining the same six motion degrees of freedom and the original gripper command. The conversion below preserves the physical motion represented before scale normalization and tokenization.


\subsection{Conversion between raw and DSD representations}

Given a raw action $a_t\in\mathcal{A}_{\mathrm{raw}}$, we first
form its translation increment and relative rotation:
\begin{equation}
    \begin{aligned}
        p_t &= (\Delta x_t,\Delta y_t,\Delta z_t)^\top,\\
        R_t &= R_z(\Delta yaw_t)R_y(\Delta p_t)R_x(\Delta r_t).
    \end{aligned}
    \label{eq:dsd-motion}
\end{equation}
For translation, the scale is the Euclidean displacement magnitude,
and the direction is its unit vector:
\begin{equation}
    s_t^{\mathrm{pos}}=\|p_t\|_2,\qquad
    n_t^{\mathrm{pos}}=
    \begin{cases}
        p_t/s_t^{\mathrm{pos}}, & s_t^{\mathrm{pos}}>0,\\
        e_3, & s_t^{\mathrm{pos}}=0,
    \end{cases}
    \label{eq:dsd-translation}
\end{equation}
where $e_3=(0,0,1)^\top$ is a fixed default direction. Its choice has
no effect on the represented action when the scale is zero.

For rotation, let $\omega_t\in\mathbb{R}^3$ be an axis--angle rotation
vector satisfying
\begin{equation}
    R_t=\exp\!\left([\omega_t]_\times\right),\qquad
    \|\omega_t\|_2\leq\pi,
    \label{eq:dsd-rotation-vector}
\end{equation}
where $[v]_\times$ denotes the skew-symmetric matrix satisfying
$[v]_\times x=v\times x$. The rotation angle and unit axis are
\begin{equation}
    s_t^{\mathrm{ori}}=\|\omega_t\|_2,\qquad
    n_t^{\mathrm{ori}}=
    \begin{cases}
        \omega_t/s_t^{\mathrm{ori}}, & s_t^{\mathrm{ori}}>0,\\
        e_3, & s_t^{\mathrm{ori}}=0.
    \end{cases}
    \label{eq:dsd-rotation}
\end{equation}
We use a numerically stable matrix-to-axis--angle conversion near zero
and $\pi$. At angle $\pi$, the equivalent axes $n$ and $-n$ are resolved
using a fixed sign convention.
The resulting scales and unit vectors, together with $g_t$, form
$\tilde{a}_t$ in Eq.~\eqref{eq:dsd-representation}.

Conversely, a DSD action specifies the motion through
\begin{equation}
    \begin{aligned}
        p_t &= s_t^{\mathrm{pos}}n_t^{\mathrm{pos}},\\
        R_t &= \exp\!\left(
            s_t^{\mathrm{ori}}[n_t^{\mathrm{ori}}]_\times\right).
    \end{aligned}
    \label{eq:dsd-reconstruction}
\end{equation}
Converting $R_t$ to roll--pitch--yaw using the original convention
and retaining $g_t$ recovers a raw action representing the same
physical command.

\begin{figure*}[t]
    \centering
    \includegraphics[width=\textwidth]{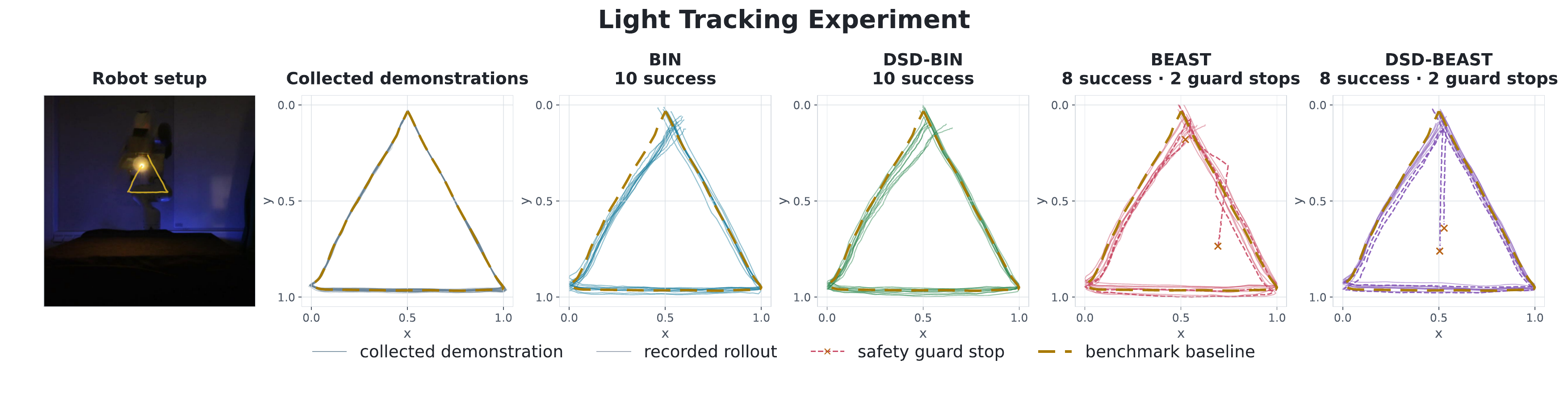}    
    \caption{Light-tracking setup and normalized trajectory comparisons. From left to right: robot setup, demonstrations collected at five execution speeds, and 10 evaluation rollouts each for BIN, DSD-BIN, BEAST, and DSD-BEAST. Gold dashed curves denote the reference trajectory, and cross markers indicate safety-guard terminations. BEAST and DSD-BEAST each have two terminated rollouts.}
    \label{fig:light}
\end{figure*}


Before tokenization, we normalize the translation magnitude and rotation angle to $[-1,1]$. We normalize actual motion magnitudes, accounting for multiplicative factors in stored actions, using shared parameters during co-training to support potential zero-shot transfer and target-specific parameters during fine-tuning. The directional components already lie within $[-1,1]$ and require no dataset-specific normalization. For DSD-BIN, each normalized scale and each directional component is discretized into 256 equal-width bins over this interval, while the gripper command retains its original encoding. At inference, the decoded scale channels are denormalized before reconstructing physical motion. The same continuous representation can also be supplied to compression-based tokenizers such as BEAST.

DSD defines the continuous representation supplied to the tokenizer and can therefore be combined with existing tokenizers operating on normalized DSD action chunks.

\subsection{Stability: masking direction loss at small scale}

When translation magnitude or rotation angle approaches zero, the corresponding direction or axis becomes sensitive to noise. Small perturbations can therefore produce substantially different directional targets for physically negligible motions. Applying the direction-token loss at full weight on these steps can introduce unreliable supervision.

We down-weight the corresponding direction-token loss when the scale falls below a threshold $\tau$, retaining a minimum weight of $w_{\min}=0.1$. This reduces the influence of noisy directional targets while preserving a nonzero learning signal. At small motion magnitudes, directional errors have a limited effect on the reconstructed displacement and do not affect execution when the resulting commands remain within the downstream controller's dead zone.


\section{Experiments}
\label{sec:experiments}

We evaluate the following questions motivated by the problems identified in Sec.~\ref{sec:problem}:
\begin{itemize}
\item \textbf{A1} -- How does a DSD policy perform relative to the raw action representation;
\item \textbf{A2} -- How compatible is DSD with different tokenization schemes;
\item \textbf{A3} -- What is the performance impact of DSD on heterogeneous co-training;
\item \textbf{A4} -- How does DSD perform when fine-tuning representation-matched pretrained checkpoints.
\end{itemize}
As detailed in the subsections below, the light-tracking experiment evaluates A1 and A2 under demonstration-speed variation, connecting the evaluation to Problem~1. LIBERO evaluates A1 and A2 on manipulation tasks. SimplerEnv evaluates A1 and A3 through single-dataset and heterogeneous co-training comparisons, motivated by Problems~2--3. Finally, real-robot manipulation evaluates A1 and A4, testing DSD's effectiveness both without robotics pretraining and after fine-tuning pretrained checkpoints.
\paragraph{Setup}
All in-repository policies are trained with Florence-2-Base with 0.23B parameters~\cite{florence}. The policies predict a 20-step action chunk for the end-effector pose change with parallel decoding: decoder self-attention is bidirectional, and one discrete distribution is produced per position. Each action dimension uses $N_b=256$ bins. The vision backbone (DaViT) and language backbone (BART encoder-decoder) are initialized from the public Florence-2-base checkpoint; the full model is fine-tuned end-to-end. 

We compare four in-repository representation and tokenizer combinations -- \textbf{BIN} (OpenVLA-style uniform scalar binning~\cite{kim2024openvla} extended to 20-step action chunks), \textbf{DSD-BIN} (DSD with BIN), \textbf{BEAST}~\cite{zhou2025beast}, and \textbf{DSD-BEAST} (DSD with BEAST). For the real-robot tasks, we additionally evaluate the pretrained-checkpoint variants \textbf{BIN-P} and \textbf{DSD-BIN-P}. BEAST represents each 20-step action chunk using 10 B-spline basis functions.

\subsection{Motion Tracking with Speed Varying Demonstrations}
To test whether DSD mitigates the sensitivity to demonstration speed described in Problem~1, we conduct a controlled path-tracking experiment using demonstrations collected at different speeds. A fixed camera tracks a light marker mounted on the robot's end-effector. A scripted controller collects 50 demonstrations (5 velocities $\times$ 10 repetitions), and their pointwise mean after arc-length resampling defines the reference trajectory. We compare BIN, DSD-BIN, BEAST and DSD-BEAST, each with 10 evaluation rollouts.

For evaluation, each executed path and the reference are independently centered at their bounding-box centers, uniformly scaled by their maximum bounding-box spans, and resampled to 256 equally spaced arc-length points. Let $X=(x_i)$ and $R=(r_j)$
denote the resulting executed and reference paths. We compute an endpoint-constrained, monotone dynamic time warping (DTW) alignment using Euclidean local distances:
\begin{equation}
\begin{aligned}
\pi^\star
&= \operatorname*{arg\,min}_{\pi \in \Pi}
   \sum_{(i,j)\in\pi} \|x_i-r_j\|_2, \\
d_{\mathrm{DTW}}
&= \frac{1}{|\pi^\star|}
   \sum_{(i,j)\in\pi^\star} \|x_i-r_j\|_2,
\end{aligned}
\end{equation}
where $\Pi$ denotes the set of valid DTW alignments. This distance measures agreement in normalized path shape and traversal order.

We report $d_{\mathrm{DTW}}$ as the primary metric and additionally
provide a demonstration-calibrated score:
\begin{equation}
\mathrm{Score}
= 100\,\operatorname{clip}\!\left(
\frac{0.05-d_{\mathrm{DTW}}}{0.05-d_{\mathrm{demo}}},
0,1\right).
\end{equation}
Here, $d_{\mathrm{demo}}=0.005768$ is the mean DTW distance obtained
by comparing each demonstration with the mean of the
other 49, using the same preprocessing. A DTW distance at or below this demonstration-level error receives 100 points, whereas a distance of 0.05 or greater receives zero.

\begin{table}[h]
\centering
\caption{Average and best tracking performance over 10 rollouts. Best denotes minimum DTW
distance or maximum score.}
\label{tab:triangle}
\begin{tabular}{lcc}
\toprule
Method & Average & Best \\
\midrule
\multicolumn{3}{c}{DTW distance ($\times 10^{-2}$) $\downarrow$} \\
\midrule
BIN       & 1.997 & 1.414 \\
DSD-BIN   & \textbf{1.588} & \textbf{0.781} \\
BEAST     & 2.743 & 1.222 \\
DSD-BEAST & 2.465 & 0.947 \\
\midrule
\multicolumn{3}{c}{Score with 0.05 cutoff $\uparrow$} \\
\midrule
BIN     & 67.90 & 81.08 \\
DSD-BIN & \textbf{77.13} & \textbf{95.38} \\
\quad $\Delta$ (vs.\ BIN) & +9.23 & +14.30 \\
BEAST   & 57.02 & 85.41 \\
DSD-BEAST &  68.25 & 91.62 \\
\quad $\Delta$ (vs.\ BEAST) & +11.23 & +6.21 \\
\bottomrule
\end{tabular}
\end{table}

Table~\ref{tab:triangle} reports average and best-case performance
over all 10 rollouts per method, including failed rollouts. Evaluation stops automatically when the light marker returns to within 2.5 cm of its starting point from below. Two BEAST and two DSD-BEAST rollouts were terminated by the maximum-rotation safety guard. DSD-BIN reduces mean DTW distance by 20.5\% relative to BIN, while DSD-BEAST reduces it by 10.1\% relative to BEAST. DSD also increases the average score by 9.23 points with BIN and by 11.23 points with BEAST; the corresponding best-score improvements are 14.30 and 6.21 points. These results answer A1 by showing that DSD better preserves path geometry when demonstrations vary in speed, directly addressing Problem~1. The gains with both tokenizers also provide evidence favouring DSD for A2.

\subsection{Simulation: LIBERO}
We evaluate on LIBERO's Spatial, Object, Goal, and Long suites~\cite{liu2023libero}, with 500 trials per suite. Table~\ref{tab:libero} compares BIN and BEAST, each with and without DSD, alongside pretrained $\pi_0$, $\pi_{0.5}$, and $\pi$-FAST baselines fine-tuned on LIBERO. DSD-BIN improves over BIN on three of four suites, increasing average success by 3.8 percentage points. DSD-BEAST improves over BEAST on all four suites, with an average gain of 3.9 percentage points, and achieves the highest Long-suite success among the compared methods (88.6\%). Without robotics pretraining, DSD-BIN outperforms pretrained $\pi$-FAST in average success (92.3\% vs.\ 85.6\%) and matches pretrained $\pi_0$ (92.3\%), which uses flow matching for continuous action prediction. These results further support DSD's effectiveness as an action representation (A1) and its compatibility with structurally different tokenizers without tokenizer-specific modifications (A2).

\begin{table}[h]
\centering
\caption{LIBERO success rates (\%). ``P'' indicates robotics pretraining. The highest success rate in each column is shown in bold.}
\label{tab:libero}
\setlength{\tabcolsep}{3.1pt}
\renewcommand{\arraystretch}{1.06}
\resizebox{\columnwidth}{!}{%
\begin{tabular}{lcrrrrr}
\toprule
Method & P & Spatial & Object & Goal & Long & Avg.\\
\midrule
BIN & -- & 83.4 & 93.4 & 91.8 & 85.2 & 88.5\\
DSD--BIN & -- & {93.6} & 96.8 & {95.0} & 83.8 & {92.3}\\
BEAST & -- & 88.4 & 96.4 & 88.6 & 78.4 & 88.0\\
DSD--BEAST & -- & 92.4 & {97.2} & 89.4 & \textbf{88.6} & 91.9\\
\midrule
$\pi_{0.5}$ & yes & \textbf{98.0} & 97.6 & \textbf{96.6} & 82.0 & \textbf{93.6}\\
$\pi$-FAST & yes & 96.4 & \textbf{98.0} & 87.8 & 60.0 & 85.6\\
$\pi_0$ & yes & 96.4 & 97.4 & 95.4 & 80.0 & 92.3\\
\bottomrule
\end{tabular}}
\end{table}

\subsection{SimplerEnv}

\begin{figure*}[t]
    \centering
    \includegraphics[width=\textwidth]{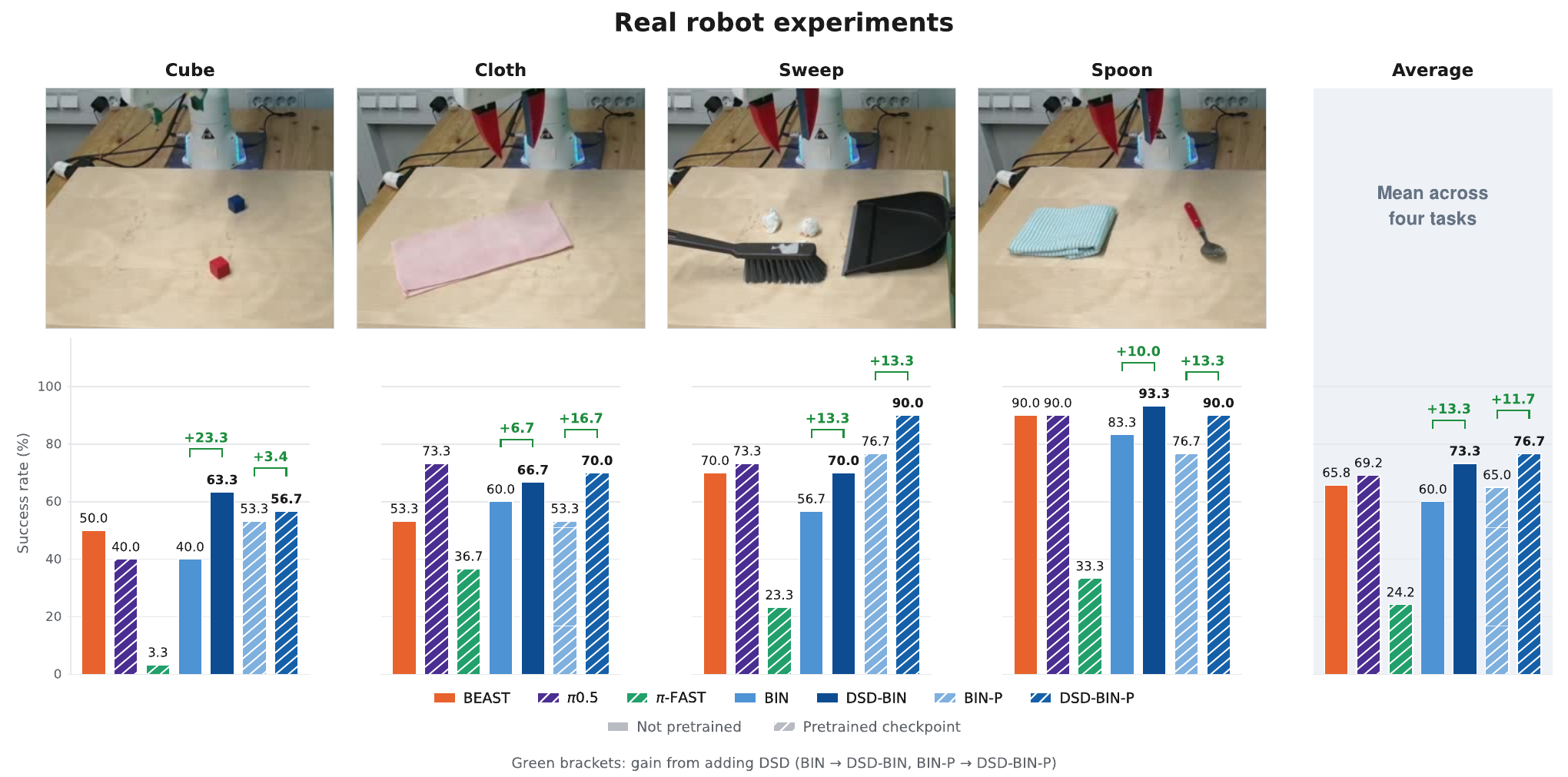}  
    \caption{Real-robot manipulation results. Top: task setups for cube stacking, cloth folding, tissue sweeping, and spoon replacement. Bottom: success rates over 30 trials per method per task, with the four-task average shown on the right. Hatched bars indicate robotics pretraining, and green brackets show gains from adding DSD in percentage points.}    
    \label{fig:realrobot}
\end{figure*}

To evaluate DSD under the heterogeneous training conditions motivating Problems~2--3 (A3), we fine-tune BIN and DSD-BIN on either bridgeV2 alone or a five-dataset OXE mixture comprising bridgeV2, Berkeley-autolab-UR5, stanford-hydra, taco-play, and viola. The mixture spans three robot embodiments: WidowX, UR5, and Franka. We evaluate Bridge tasks on SimplerEnv~\cite{li2024simplerenv} using four tasks with 75 matched trials per task.

As shown in Table~\ref{tab:simplerenv}, DSD-BIN exceeds BIN by 0.7 percentage points under Bridge-only training. Adding the four datasets increases DSD-BIN's overall success from 40.7\% to 43.3\%, a gain of 2.7 percentage points, while reducing BIN's success from 40.0\% to 33.0\%. Under co-training, DSD-BIN therefore outperforms BIN by 10.3 percentage points (paired, task-stratified bootstrap 95\% CI: 5.0--15.7 points; Holm-adjusted exact McNemar $p=0.003$), based on 300 matched evaluation trials.

The benefits vary across tasks: co-training improves DSD-BIN's carrot and eggplant scores but reduces its spoon and stack scores. Despite using robot data from only the five-dataset mixture, DSD-BIN-mix achieves the highest overall success rate (43.3\%) among all methods in Table~\ref{tab:simplerenv}, including finetuned baselines with large-scale robotics pretraining. These results support DSD in answering A1, as DSD-BIN achieves higher overall success than BIN under both single-dataset and mixed-dataset training. They also answer A3 positively: explicitly separating direction and scale enables an aggregate benefit from additional heterogeneous data, whereas BIN degrades under the same training mixture.

\begin{table}[h]
\centering
\caption{SimplerEnv success rate (\%). ``-bridge'' = fine-tuned on bridgeV2 only; ``-mix'' = co-trained on the 5-dataset mixture.}
\label{tab:simplerenv}
\resizebox{\columnwidth}{!}{%
\begin{tabular}{lccccc}
\toprule
Method & Carrot & Spoon & Stack & Eggplant & Overall \\
\midrule
RT-1-X            & 10.7 & 4.0  & 0.0  & 0.0  & 3.7 \\
Octo-base         & 6.7  & 8.0  & 0.0  & 41.3 & 14.0 \\
Octo-small        & 5.3  & 34.7 & 2.7  & 53.3 & 24.0 \\
OpenVLA           & 0.0  & 0.0  & 0.0  & 0.0  & 0.0 \\
$\pi$-FAST-ft     & 0.0  & 0.0  & 0.0  & 9.3  & 2.3 \\
$\pi_0$-ft        & 8.0  & 6.7  & 6.7  & 14.7 & 9.0 \\
$\pi_{0.5}$-ft    & \textbf{69.3} & 34.7 & 14.7 & 40.0 & 39.7 \\
SpatialVLA        & 22.7 & 13.3 & \textbf{16.0} & 94.7 & 36.7 \\
\midrule
BIN-bridge        & 20.0 & 40.0 & 4.0  & \textbf{96.0} & 40.0 \\
BIN-mix           & 12.0 & \textbf{49.3} & 12.0 & 58.7 & 33.0 \\
DSD-BIN-bridge    & 45.3 & 41.3 & 9.3  & 66.7 & 40.7 \\
DSD-BIN-mix       & 49.3 & 40.0 & 4.0  & 80.0 & \textbf{43.3} \\
\bottomrule
\end{tabular}}
\end{table}

\subsection{Real-robot experiments}
We evaluate four real-robot tasks---cube stacking, cloth folding, tissue sweeping, and spoon replacement---with controlled initial conditions and 30 trials per method per task. We compare BEAST, $\pi_{0.5}$-DROID, $\pi$-FAST-DROID, BIN, DSD-BIN, BIN-P, and DSD-BIN-P. BIN-P and DSD-BIN-P are fine-tuned from the corresponding BIN-mix and DSD-BIN-mix checkpoints obtained through five-dataset OXE co-training in the SimplerEnv experiments. The pretraining data include related tasks or motion patterns.

Cube stacking succeeds when the red cube remains stably on the blue cube after release; cloth folding requires one side of the cloth to be folded onto the other, with contact covering more than two-thirds of the folded side; tissue sweeping requires all pieces to lie inside the dustpan; and spoon replacement requires the spoon to reach the target cloth. These tasks cover various interactions, including contact-rich manipulation and behaviors beyond pick-and-place.

Figure~\ref{fig:realrobot} reports per-task success rates. DSD-BIN outperforms BIN on every task (Cube +23.3, Cloth +6.7, Sweep +13.3, Spoon +10.0), and the pretrained DSD-BIN-P likewise outperforms BIN-P on every task as well (Cube +3.4, Cloth +16.7, Sweep +13.3, Spoon +13.3). Averaged over all four tasks, DSD increases success rate by +13.3 points without pretraining (BIN 60.0\%$\to$DSD-BIN 73.3\%) and +11.7 points with pretraining (BIN-P 65.0\%$\to$DSD-BIN-P 76.7\%). These results support DSD's effectiveness on real robots and its continued benefits when fine-tuning representation-matched pretrained checkpoints (A4).

\section{Discussion}
\label{sec:discussion}

Our simulation and real-robot results suggest that separating direction and scale improves learning from demonstrations that vary in speed or motion scale. In raw coordinate-based action representations, execution-speed variation can obscure shared geometric patterns, while dataset-specific normalization can produce inconsistent token targets and distort motion direction during transfer. DSD exposes this geometric structure explicitly, reducing the need to infer it from token identities alone. The mixed-dataset results are consistent with this motivation: adding four datasets improves DSD-BIN's overall performance but reduces BIN's.

Several limitations warrant further investigation. First, computational constraints restrict our mixed-dataset evaluation to five OXE datasets, leaving DSD's effectiveness on larger mixtures, including the full OXE collection, untested. Second, gains on SimplerEnv vary across tasks, motivating further analysis of the task characteristics and failure modes underlying this variation. Nevertheless, by retaining magnitudes in dedicated scale channels, the decomposition preserves the original motion information before quantization, even for tasks without a ``geometry path invariance'' structure. Third, the threshold used to stabilize direction learning may require tuning across training stages. In our experiments, fine-tuning thresholds ($0.05$--$0.3\%$ of the full scale range) are generally smaller than the pretraining threshold ($0.5\%$).

Our tokenizer evaluation is limited to BIN and BEAST. Future work should examine DSD's compatibility with other tokenizers and whether tokenization schemes tailored to its structure offer further gains. Our evaluations also cover only Franka in the real world and Franka and WidowX in simulation. Broader embodiment coverage and extensions to bimanual manipulation are therefore important practical directions. Finally, motion reconstructed from monocular egocentric demonstrations may have uncertain metric scale. DSD preserves translation direction under uniform positive rescaling, potentially providing consistent directional supervision while allowing motion magnitudes to be calibrated separately. Future work will evaluate whether this property improves learning from egocentric demonstrations.

\section{Conclusion}
\label{sec:conclusion}
 
We examined three problems of raw pose-increment representations: sensitivity to execution speed, inconsistent token targets across datasets, and motion-direction distortion under mismatched normalization. DSD separates translation into direction and magnitude and relative rotation into axis and angle before normalization, while retaining the gripper command. Fixed directional bounds provide a common numerical convention across datasets, and the analytic representation can be combined with existing tokenizers. Simulation and real-robot experiments show improvements in average success and normalized path fidelity, with benefits also observed after fine-tuning representation-matched pretrained checkpoints. In the tested five-dataset mixture, DSD-BIN benefits from additional heterogeneous data while BIN's overall performance declines. These findings support DSD as an effective action representation for discrete-token VLAs and motivate further controlled studies across larger dataset mixtures, additional tokenizers, and broader embodiments.

\section*{Acknowledgments}

OpenAI Codex was used for generating the figures. The authors manually revised the figures and verified all labels, plotted data and numerical values.


\bibliographystyle{IEEEtran}

\bibliography{references}  

\end{document}